\documentclass{article}

\usepackage{PRIMEarxiv}

\usepackage{amssymb}

\usepackage{amsmath}

\usepackage{graphicx}
\usepackage{subcaption}
\PassOptionsToPackage{hyphens}{url}\usepackage{hyperref}

\usepackage{algorithm}
\usepackage{algpseudocode}
\usepackage[table]{xcolor}

\title{Analysing Environmental Efficiency in AI for X-Ray Diagnosis
\thanks{\textit{\underline{Published in Journal of AI under CC BY 4.0. Available at }}: 
\textbf{DOI:
10.61969/jai.1838517}} 
}

\author{
  Liam Kearns \\
  AuraQ \\
  Malvern \\
  Worcestershire \\
  UK \\
  \texttt{liamkearnsy@gmail.com} \\
}

\begin{document}
\maketitle

\begin{abstract}
The integration of AI tools into medical applications has aimed to improve the efficiency of diagnosis. The emergence of large language models (LLMs), such as ChatGPT and Claude, has expanded this integration even further despite a concern for their environmental impact. Because of LLM versatility and ease of use through APIs, these larger models are often utilised even though smaller, custom models can be used instead. In this paper, LLMs and small discriminative models are integrated into a Mendix application to detect Covid-19 in chest X-rays. These discriminative models are also used to provide knowledge bases for LLMs to improve accuracy. This provides a benchmark study of 14 different model configurations for comparison of diagnostic accuracy and environmental impact. The findings indicated that while smaller models reduced the carbon footprint of the application, the output was biased towards a positive diagnosis and the output probabilities were lacking confidence. Meanwhile, restricting LLMs to only give probabilistic output caused poor performance in both accuracy and carbon footprint, demonstrating the risk of using LLMs as a universal AI solution. While using the smaller LLM GPT-4.1-Nano reduced the carbon footprint by 94.2\% compared to the larger models, this was still disproportionate to the discriminative models; the most efficient solution was the Covid-Net model. Although it had a larger carbon footprint than other small models, its carbon footprint was 99.9\% less than when using GPT-4.5-Preview, whilst achieving an accuracy of 95.5\%, the highest of all models examined. This paper contributes to knowledge by comparing generative and discriminative models in Covid-19 detection as well as highlighting the environmental risk of using generative tools for classification tasks.

\end{abstract}



\section{Introduction}
\label{sec:intro}

The rapid commercialisation of large language models (LLMs) has significantly expanded the use of artificial intelligence (AI) across industries. As these models have grown in their capabilities, their size have increased, requiring a greater level of computational resources. With GPT-4 consisting of more than 1.7 trillion parameters \cite{Hadi2023}, the size of these models can be in the terabyte range. This has created a reliance on third-party cloud infrastructure, raising concerns about energy consumption and environmental sustainability.

The medical field has been significantly influenced by advances in AI. This can be observed through the use of AI for the detection of Covid-19 from chest X-rays  \cite{Tzeng2023}. Furthermore, commercialised LLMs, including ChatGPT and Claude, have been proposed as tools to assist with medical diagnoses through text analysis \cite{Sorin2023, Pedram2024}, as well as image analysis \cite{Nguyen2024, Aydin2025}. However, an overreliance on third-party models is not a universal solution. Sharing data with external providers can cause security risks and infringe on regulations if malpractice risks patient harm \cite{Shumway2024, ChenAI2025}. LLMs can show false overconfidence in performing straightforward, case-specific tasks \cite{Marianna2025}. Additionally, using API calls to an external host can result in higher latency and costs, risking an increased carbon footprint. This is a prominent issue in the medical field, with medical imaging already emitting high levels of greenhouse gases \cite{Doo2024}. As AI assistance in this field increases, concerns about the carbon emissions caused by using third-party LLMs will become a prominent issue.

An alternative approach is to deploy models within the environment of an application. Although this reduces latency and gives greater control over data security, resource consumption becomes a major concern and limitation. It is impractical to deploy commercial-sized LLMs in application environments; therefore, local models must be designed and developed to consume less memory. Classification models, including ResNet and VGG, which are significantly smaller than LLMs, have demonstrated high specificity in detecting Covid-19 from X-rays \cite{Tzeng2023}, highlighting smaller models have practicability in medical applications. However, reducing model sizes typically means narrowing the scope and accuracy. This trade-off is not suitable for the medical field, as inaccurate disease detection can result in delayed diagnosis and treatment \cite{Noguchi2023}. Consequently, reducing model memory to reduce carbon footprints should not compromise model accuracy.

This paper investigates the diagnostic accuracy and carbon footprint across generative LLMs and smaller discriminative models in an AI-assisted medical application. Specifically, models are compared to analyse whether LLMs are a sustainable solution for narrow medical classification tasks, or if smaller discriminative models provide environmental efficiency without compromising diagnostic accuracy. In the context of this paper, the terms 'small models' and 'local models' are used interchangeably. These terms refer to the discriminative models deployed in the application environment discussed in Section \ref{sec:method}.

\begin{figure} [htbp]
    \centering
    \includegraphics[height=2.5in]{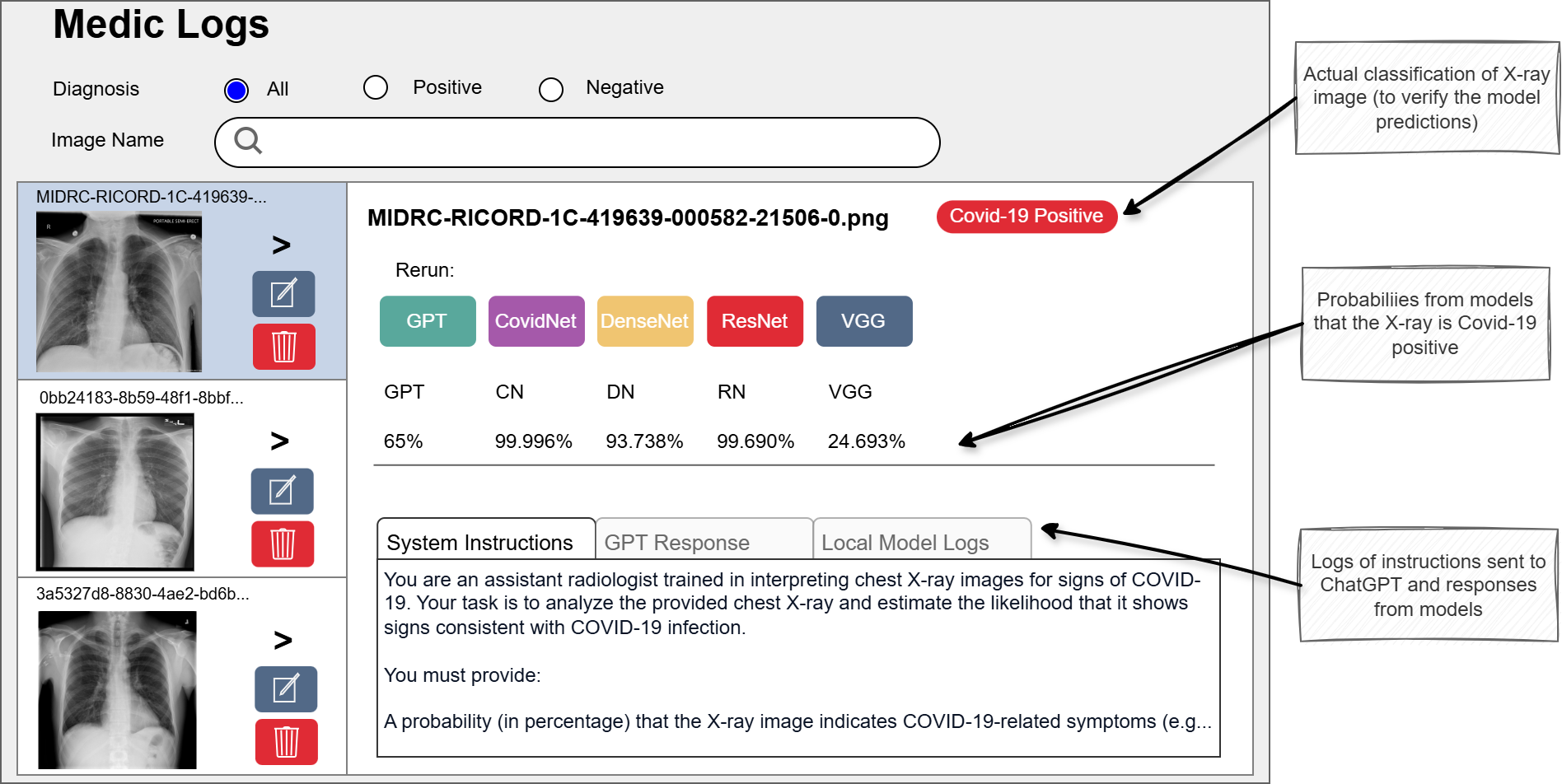}
    \caption{Application using machine learning models to diagnose Covid-19 in chest X-rays.}
    \label{fig:scope}
\end{figure}

The scope of this paper is focused on using AI models in a medical context for the detection of Covid-19 from chest X-rays. As illustrated in Figure \ref{fig:scope}, the application used in this paper uses both externally hosted LLMs and local models to determine the probability that an X-ray shows signs of Covid-19. Given the implementation of discriminative models as local models, a classification use case was required to train these models. Although generative models have a larger scope of capabilities than just outputting probabilities, the focus on a single disease facilitates straightforward comparison between models, with the model outputs determining whether an X-ray shows signs of the presence of Covid-19. Although results may not generalise to multi-class diagnostic settings or other diseases, this focused scope allows clear evaluation of environmental and performance trade-offs in a controlled scenario.

The structure of the paper is as follows. Section \ref{sec:related} highlights state-of-the-art models for X-ray analysis when using LLMs and smaller discriminative models, as well as their environmental impact. Next, Section \ref{sec:method} outlines how LLMs and local models will be integrated into a medical application to compare their performance in detecting Covid-19 from X-ray images. Section \ref{sec:result} discusses these results, evaluating their suitability to be used in medical diagnosis before identifying limitations within this study.

\subsection{Contributions}
This paper focuses on a comparative analysis of LLMs and discriminative models, while simultaneously quantifying their environmental impact within a medical application. The paper's major contributions are as follows:

\begin{itemize}
    \item Demonstrating major accuracy and carbon footprint concerns when using LLMs for probabilistic outputs to classify diseases present in X-rays.
    \item Highlighting both diagnosis accuracy and carbon footprint can be reduced by using local discriminative models, challenging the assumption that larger, general-purpose models are universally advantageous
    \item Showing that focusing on reducing carbon footprint over model accuracy can diminish the confidence and accuracy of smaller model outputs.
    \item Identifying that knowledge bases for LLMs can increase detection accuracy, but has a varied impact on carbon footprint. 

\end{itemize}

\section{Related works}
\label{sec:related}

\subsection{Large language models}
The medical field has recently been introduced to the capabilities of LLMs. Implemented models have shown an accuracy of more than 70\% when verified by medical professionals \cite{Thawakar2025}. Notably, ChatGPT has shown accurate identification across various diseases \cite{Ostrovsky2025}. This highlights the versatility of LLMs, in which models can be used to detect and evaluate multiple pathologies.

Although LLMs provide robustness through generalisation, there are risks on the integrity of their outputs. Censorship or bias can occur unintentionally when models are trained on datasets containing censored information \cite{Ahmed2024}. More malicious threats to the integrity of LLM outputs can be caused by prompt manipulation. Although implementing semantic censorship can mitigate malicious prompts, they can be split up and re-written to disguise malicious intent \cite{Glukhov2023}. Prompt injection is also a vulnerability for some LLMs, where malicious users can make predefined prompts redundant or leak prompts that are hidden from end users \cite{Liu2024}. In medical contexts, bypassing LLM safety mechanisms can be dangerous to the health of patients \cite{ChenAI2025}. These vulnerabilities can be exploited if end user inputs are used directly in the prompt subsequently sent to LLMs. Therefore, it is important to have a level of doubt in LLM outputs due to the risks surrounding their training and inference.

Training LLMs for medical purposes is a challenge due to ethical concerns. Medical information is highly sensitive, so mismanagement of data permission can result in serious legal action. To ensure compliance, the algorithms and training procedures of LLMs must be independently reviewed \cite{Shumway2024}. With commercialised LLM algorithms often not open source, peer-reviewing their compliance with sensitive patient information is not always possible. This results in less data available for LLMs to train on. Consequently, models can identify unhealthy medical images, but can struggle with the precise classification of specific abnormalities \cite{Aydin2025}. Additionally, diagnosing less common conditions results in low accuracy \cite{Ostrovsky2025}. Without a secure way to train on medical data without compromising patient safety, obtaining accurate medical results from third-party models will continue to be a challenge.

An alternative to fine-tuning LLMs is to use a knowledge base to provide the model with additional information about the medical question being asked. Retrieval augmented generation (RAG) has been used to reduce hallucination and improve consistency in generated medical reports \cite{Ranjit2023}. RAGs using cosine similarity to retrieve the most relevant information have seen medical analysis with ChatGPT models increase more than 20\% \cite{Tayebi2025}. Knowledge bases can also be used to retrieve information from trusted sources, resulting in more reliable advice \cite{ZhaoGrAI2024}. This demonstrates an alternative method of providing LLMs with additional information without the need for fine-tuning.

Attempting to automate clinical actions through LLMs poses safety concerns. Although existing studies show promising results in high accuracy of LLMs in X-ray analysis, disruption and damage can occur when LLMs output biased or factually incorrect information \cite{Pedram2024}. LLM performance can improve when the context of a positive diagnosis and recommended next steps are provided within the prompt, but models can overlook important information due to their attention mechanism \cite{Sorin2023}. As it is impractical to expect consistent and accurate outputs from a model, it is important to implement safeguards to prevent unfiltered communication between the model and patients to ensure data security and patient safety.

LLMs can provide explanations for their answers, which can be beneficial when defending clinical decisions. However, these explanations are not guaranteed to be factual. Reasoning models have struggled to be accurate in scenarios where the output format is expected to be consistent \cite{Marianna2025}. In medical scopes, ChatGPT has shown confidence in the reasoning behind its results in relation to X-ray analysis, even when its diagnosis is incorrect \cite{Bera2024}. Although other models such as Claude-3.5 Sonnet have shown more consistent results in medical tasks, low sensitivity values highlight that confusion is still prevalent \cite{Nguyen2024}. Improvements can be made to minimise LLM hallucinations by verifying quantitative measurements provided in the model outputs \cite{Heiman2024}. However, these measures rectify errors generated by LLMs rather than fine-tuning LLMs to avoid these hallucinations from occurring. So, while steps can be taken to rectify incorrect outputs, the risks of hallucinations in model output still pose a threat to medical diagnoses.

\subsection{Small models}

Although LLMs have seen increased focus and improved accuracy, smaller models have not been made obsolete. Notably, LLMs struggle with classification tasks, especially when the correct label is not provided in the input \cite{Hanzi2024}. So, while the versatility of LLMs allows classification tasks to be performed, comprehension of the task cannot be guaranteed. Moreover, when resources are limited and a project has a narrow scope, the use of small models has been shown to be advantageous \cite{Lihu2024}. Where training is restricted, smaller models have been shown to outperform larger models \cite{Hassid2024}. This enhanced performance with reduced resources leads to increased efficiency in training and acquiring outputs from smaller models. In medical scopes, this is beneficial when substantial data on a novel disease is unavailable or when there is insufficient funding to train models on energy-intensive equipment.

Neural networks have been shown to provide accurate classifications in medical fields. The VGG model has achieved an accuracy of more than 92\% in detecting pneumonia on chest X-rays \cite{Sharma2022}. These high accuracies are partially obtained by restricting models through classification, whereby outcomes are grouped. Discriminative models of this nature require labelled data for training. Expanding from binary classification to multi-classification (e.g., classifying X-rays as normal, Covid-19, or pneumonia) can reduce accuracy to 82\% \cite{Bharati2021}. So, while neural networks can achieve high accuracies through classification, this also narrows their scope, making their use case specialised.

Covid-Net is a neural network design that focuses on detecting Covid-19 from chest X-rays. The initial version of the model achieved an accuracy of 93.3\% \cite{WangLinda2020}. Over time, the open source initiative has increased, resulting in the Covidx CXR-3 dataset, allowing Covid-Net models to reach accuracies of over 96\% \cite{Pavlova2022}. Existing model architecture has also been used to accurately detect Covid-19 in X-ray images. A fine-tuned DenseNet model achieved 92\% accuracy in detecting Covid-19 despite its reduced complexity compared to other models \cite{Albahli2021}. This ability to fine-tune models for medical scenarios highlights the robustness of smaller models. Although not as accurate as models like Covid-Net, which are built specifically for Covid-19 detection, the reduced complexity results in less memory usage, making it suitable for resource-constrained environments.

\subsection{Environmental efficiency of models}

Hosting machine learning models on third-party cloud computing servers can help reduce their carbon footprint through shared resources. With large cloud providers that allow AI models to be hosted in low carbon countries in efficient data centres, energy usage can be reduced to have a more positive environmental impact \cite{Lannelongue2021}. Moreover, the storage of models on the cloud allows collaboration on a model, reducing the redundancy of creating and training additional models \cite{Doo2024}. This shows that actions can be taken when models are deployed in the cloud, improving environmental sustainability.

Smaller discriminative models can be made more environmentally sustainable. Measures to reduce the carbon footprint of discriminative models have included using smaller models during times of high carbon intensity from power sources \cite{Jung2024}. However, this means accepting a lower accuracy when the carbon intensity is high. In medical contexts, more environmentally conscious models have been argued to be more competitive due to reduced training and inference times \cite{Narimani2025}. However, these models also trade accuracy for a more environmentally conscious approach. This highlights that considerations must be made when focusing on the carbon footprint of discriminative models, whereby accuracy is often negatively impacted to ensure faster model inference.

An increase in input parameters for a model does not necessarily relate to a larger carbon footprint. Factors such as power efficiency in the data centres that store the models and training time mean that models with more input parameters can have a significantly lower carbon footprint \cite{Luccioni2023}. However, language processing models risk being overparameterised, resulting in less efficient training \cite{Barbierato2024}. Therefore, while an LLM with a greater number of input parameters does not necessarily result in an increased carbon footprint, the complexity of these models requires more computationally intensive training compared to other models.

Calculating precise carbon footprints for machine learning is a challenge due to multiple distributed factors. With machine learning models often trained and deployed on external servers, obtaining exact emission data is difficult \cite{Luccioni2023}. Although newer hardware may claim to be more energy efficient, uncertainty in the manufacturing process means that a comprehensive assessment of the full environmental impact remains a challenge \cite{Davy2021}. Although precise calculations are difficult, it is evident that interacting with multimodal models, such as ChatGPT, consumes more energy than smaller task-specific models \cite{Luccioni2024}. Moreover, LLMs that require greater GPU usage result in increased carbon emissions due to more energy being consumed \cite{Jegham2025} So, while it is difficult to calculate the precise carbon footprint of models due to potential unknowns surrounding training and deployment, carbon footprints from model inference are more obtainable.

\section{Methodology}
\label{sec:method}

Machine learning provides an important tool in Covid-19 detection to improve diagnosis speed and accuracy. However, medical departments are not guaranteed an abundance of computer resources. In addition, there is an increasing concern about the carbon footprint when integrating AI models into medical applications. Although LLMs can be inferred through APIs to address local computing limitations, using such large models for binary classification may have a disproportionate environmental impact. This section outlines an application used for Covid-19 detection, in which multiple different models are tested to determine optimal accuracy and energy efficiency.

To compare the environmental efficiency of AI for X-ray diagnosis, both third-party generative and local discriminative models are used. OpenAI and Claude-3.5 models have been applied in previous studies to provide X-ray analysis \cite{Ostrovsky2025, Nguyen2024}. LLMs are hosted by various providers and have varying infrastructure, which affects their power consumption \cite{Jegham2025}. Therefore, to provide a benchmark study, models of different sizes and hardware requirements are analysed. The LLMs used in this study (GPT-4.5-Preview, GPT-4.1-Nano, o4-Mini, and Claude-3.5 Sonnet via GenAI) are multimodal large language models capable of processing both text and image inputs.

For discriminative models, Covid-Net is a model built specifically for Covid-19 detection, which makes it a suitable model to compare in this paper \cite{WangLinda2020, Pavlova2022}. The neural networks DenseNet, ResNet and VGG have shown success in X-ray tasks \cite{Sharma2022, Albahli2021}. These models are smaller in size than Covid-Net. Therefore, testing with these models alongside Covid-Net will aim to verify whether a suitable trade-off between accuracy and carbon footprint can be made, as justified in previous studies \cite{Jung2024, Narimani2025}. Using different types of models of varying size and complexity enables a comprehensive analysis of the accuracy and environmental sustainability of AI in X-ray diagnosis.

\subsection{Application}
\begin{figure}
    \centering
    \includegraphics[height=2.8in]{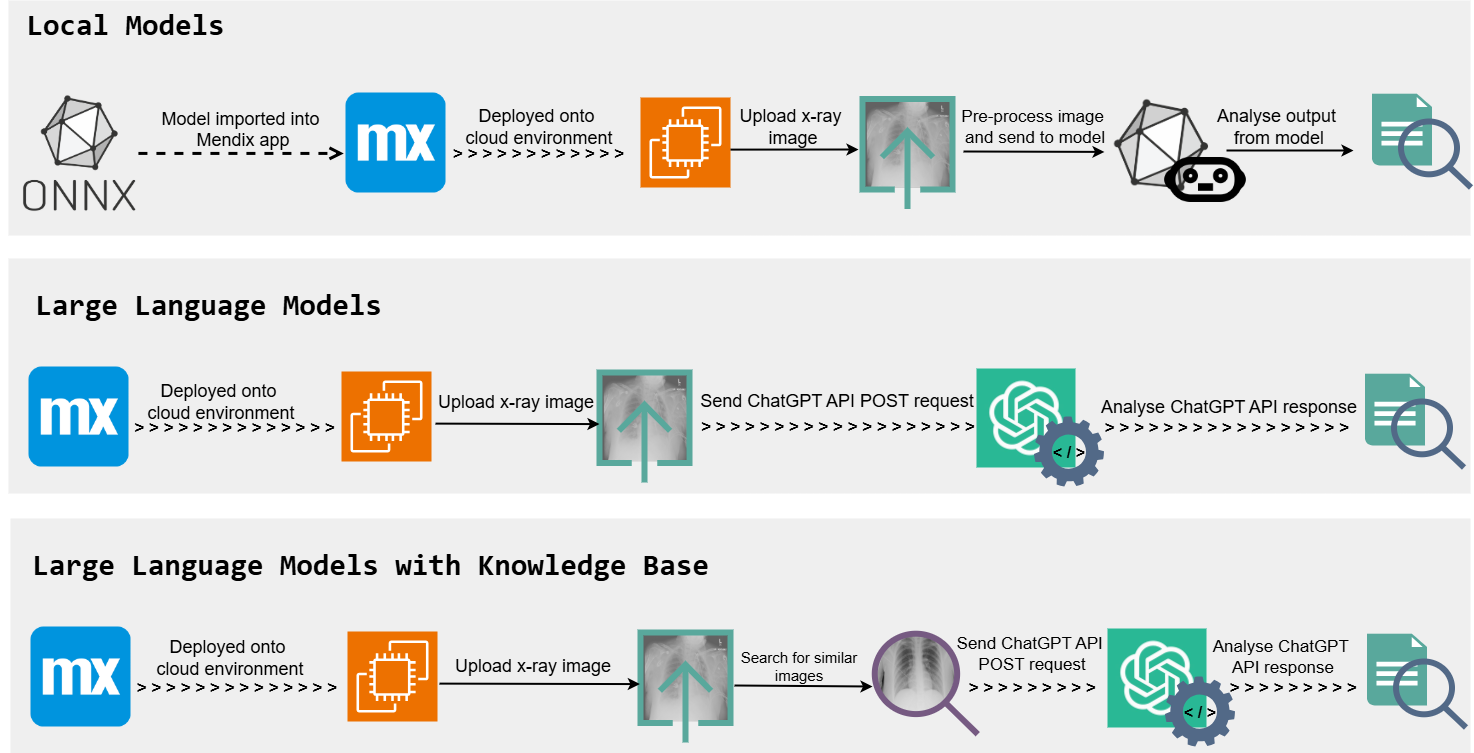}
    \caption{Deployment and use of models analysing X-ray images.}
    \label{fig:method}
\end{figure}

For this paper, the low-code platform Mendix is used to test LLMs and small models in the analysis of X-ray images. Mendix is a low-code platform that facilitates rapid application development and has seen success in developing medical applications that integrate machine learning through API capabilities \cite{Anjali2025}. Additionally, Mendix provides seamless integration of machine learning models with the ONNX format. ONNX has been observed to allow the straightforward implementation of trained models \cite{Shridhar2020}. Additionally, Mendix provides access to their own generative AI tools, GenAI, which uses Anthropic Claude-3.5 Sonnet hosted on AWS. Using this tool in conjunction with the OpenAI API within the Mendix application, multiple LLM models can be used for X-ray analysis. Figure \ref{fig:method} shows how the models will be implemented and interacted with in the Mendix application. The process of uploading and viewing the output of models is identical throughout all models. This simplifies the testing process and shows that Covid-19 detection can be performed identically regardless of the underlying model being used.

Mendix relies predominately on Java, which is not a favourable language for machine learning research. Using Java for AI algorithms has been found to have reduced CPU efficiency compared to Python \cite{Aguado2021}. However, by implementing the models in a different environment, the robustness of their accuracy will be tested outside of the research-focused environment of Python. This will help determine whether the models are suitable in scenarios where they will be deployed in applications for hypothetical real-word use. Therefore, the local models (Covid-Net, DenseNet, ResNet, and VGG) will be tested instead of using their stated accuracies from previous research to ensure that their performance is guaranteed in an unfavourable coding environment.

\subsection{Comparison}

LLM responses can provide a descriptive analysis of X-rays. However, this does not guarantee improved interpretability. Hallucination or biased responses based on input prompts can cause uncertainty regarding the accuracy of identifying Covid-19 from X-rays. This is seen in other research, where LLMs have shown confidence in erroneous outputs \cite{Bera2024, Marianna2025}. Instead, it can be argued that the output from the small models is more interpretable, since they will consistently provide a probability on signs of Covid-19 infection. This shows smaller models having improved interpretability and transparency in narrow scopes \cite{Lihu2024}. So, while LLMs can provide detailed analysis, this risks causing confusion to end users if the details are false.

To allow straightforward comparison of all models, LLMs are constrained via system instructions to return only a numeric probability of Covid-19 detection. Although LLMs can provide details on X-ray images, medical expertise is required to verify the accuracy \cite{Thawakar2025}. Furthermore, discriminative models only provide probabilities that Covid-19 symptoms are present. By enabling binary classification through restricting outputs to the probability of Covid-19, evaluation of the false-negative risk can be conducted, which is important since missed detection can result in delayed isolation and treatment \cite{Noguchi2023}; the scope of this paper is not to optimise multi-class detection, but rather analysing environmental efficiency and model accuracy, which can be achieved through this binary classification.

When analysing the time taken and carbon footprint of inferring models, control variables such as cloud hardware are important considerations. Comparing models on different servers causes uncertainty with different hardware being used \cite{Luccioni2023}. By keeping the application and local models on the same server, more consistency can be guaranteed by ensuring the resources provided throughout testing models remains as consistent as possible.

With environmentally efficient AI-assisted applications being an important consideration in this paper, the carbon footprint for running the models is an important comparison. With the Mendix cloud using Kubernetes on AWS, Equation \ref{eq:carbon} can be used to estimate the carbon footprint of running a model on the deployed application. This equation is based on estimates of power consumption of EC2 instances on AWS \cite{Davy2021}. To determine how long cloud resources are used, the time taken from uploading an X-ray to receiving probabilities is recorded. It is important to note that Kuberentes on AWS is priced every second for a minimum of 60 seconds, so the calculated carbon footprints are a hypothetical for the total runtime of the models, even if they run for less than 60 seconds.

Calculating the carbon footprint of LLMs hosted on servers is more challenging due to the variety of hardware used and unknown factors such as memory usage. However, the infrastructure specifications provided in existing research \cite{Jegham2025} can be used in Equation \ref{eq:carbon}. This research has omitted manufacturing emissions to prevent distortion when comparing models. Furthermore, assuming that manufacturing emissions are distributed across millions of daily model inferences over the lifespan of the hardware, these values should be considered negligible compared to the energy consumed through model inference.

Manufacturing emissions associated with hosting the application are not negligible, which is also where the discriminative models are hosted. Unlike LLMs, the inference of discriminative models is not shared by multiple users, so manufacturing emissions are significant and are therefore included in Table \ref{table:specs}. While manufacturing emissions will be included in the calculation of the carbon footprint produced by the Mendix application and the discriminative models, Table \ref{table:specs} shows that these emissions will be omitted when calculating the carbon footprint produced by the servers hosting the LLMs.

\begin{equation}\label{eq:carbon}
\frac{E}{1000} = (\frac{W \times P \times I}{1000} + M) \times \frac{t}{3600}
\end{equation}

\begin{equation}\label{eq:carbon_sim}
E = (\frac{W \times P \times I}{1000} + M) \times \frac{t}{3.6}
\end{equation}

\begin{align*}
  \text{where}~~  E &= \text{Carbon footprint (mgCO}_2\text{eq),}\\
  W &= \text{Instance power consumption (Watts),} \\
  P &= \text{Power usage effectiveness,} \\
  I &= \text{Electricity carbon intensity (gCO}_2\text{eq/kWh),}  \\
  M &= \text{Manufacturing emissions (gCO}_2\text{eq),} \\
  t &= \text{Time model is running (seconds)}
\end{align*}

It is important to note that the memory utilised for X-ray analysis is contingent upon the specific model used. With exact power consumption for the specific memory used being unavailable, Equation \ref{eq:carbon_mem} calculates the carbon footprint in the context of the amount of memory used in the cloud instance during X-ray analysis. When using LLMs, the size of the model ($m$) will be 0 as the model inference occurs through an API call rather than deploying the model into the application environment.

\begin{equation}\label{eq:carbon_mem}
E_m = E \times \frac{a + m}{C}
\end{equation}

\begin{align*}
  \text{where}~~  E_m &= \text{Carbon footprint per cloud memory used (mgCO}_2\text{eq/MB),}\\
  E &= \text{Carbon footprint (mgCO}_2\text{eq),}\\
  a &= \text{Application size (MB),} \\
  m &= \text{Model size (MB),} \\
  C &= \text{Cloud instance total memory (MB)}
\end{align*}

At the time of analysis, GPT-4.5-Preview was the most recent release provided by the OpenAI API. Other studies have used ChatGPT extensions for X-ray diagnosis, with the "X-Ray Interpreter" being a notable example \cite{Ostrovsky2025}. However, there was no identifiable way to use these extensions through API calls to ChatGPT through the Mendix application. Moreover, initial tests using the extension showed no discrepancy in output diagnosis probabilities compared to GPT-4.5.

\begin{table} [htbp]
\caption{Infrastructure specifications of instances.}
\begin{tabular}{ |p{2.5cm}|p{2.8cm}|p{1.8cm}|p{2.5cm}|p{2.5cm}|}
\hline Instance & Instance power consumption ($W$) & PUE & Carbon intensity ($gCO_eq/kWh$) & Manufacturing emissions ($gCO_2eq$) \\
\hline
App & 5.3 & 1.2 & 228 & 1.2  \\
\hline
GPT-4.5-Preview & 1301 & 1.12 & 353 & na \\
\hline
o4-Mini & 991 & 1.12 & 353 & na\\
\hline
GPT-4.1-Nano & 377 & 1.12 & 353 & na \\
\hline
GenAI & 1301 & 1.14 & 385 & na \\
\hline
\end{tabular}
\label{table:specs}
\end{table}

To verify whether the size of the model affected the performance in X-ray analysis, smaller variations of GPT models were also used. Therefore, analysis of the test dataset was conducted using the OpenAI API to send requests to GPT-4.5-Preview, o4-Mini, and GPT-4.1-Nano. Alongside Mendix's GenAI being analysed, this allows the benchmark study to compare five LLMs and four discriminative models.

\subsection{Dataset}

All models are evaluated using the Covidx CXR-3 dataset \cite{Pavlova2022}. Discriminative models were fine-tuned with the training distribution of the dataset. The CXR-3 test distribution labels cases as positive or negative for Covid-19. Binary classification has shown higher sensitivity and specificity compared to multi-classification models that have been trained to also detect pneumonia cases \cite{Tzeng2023}. Therefore, in addition to providing a simpler comparison, selecting these labels over multi-classification aims to improve the performance of AI models used.

The CXR-3 dataset is a cumulation of multiple open-access datasets during the Covid-19 pandemic. The training distribution contains 29,986 X-ray images from 16,648 patients, while the testing distribution includes 200 negative and 200 positive-labelled X-ray images from 378 patients \cite{Pavlova2022}. This extensive benchmark dataset contains X-rays from a variety of patients, thereby reducing the risk of overfitting to a specific demographic. With no additional patient information contained in this dataset, no additional patient-level inclusion or exclusion criteria are applied.

\subsection{Refinement with knowledge bases}

After initial analysis with the LLMs, providing additional information was necessary due to low accuracy. Knowledge bases have been shown to improve medical image analysis with LLMs \cite{ZhaoGrAI2024}. However, exposing additional sensitive information to LLMs can threaten patient safety by increasing the risk of malpractice \cite{Shumway2024}. Additionally, providing additional images increases the cost of using LLMs dramatically (providing three similar images alongside the original could hypothetically quadruple the input tokens used). Therefore, a knowledge base containing medical images was created in which a cosine similarity search retrieves the three most similar image vectors from the knowledge base. This technique has been shown to improve LLM medical analysis \cite{Tayebi2025}. The cosine similarity score and whether the image was a positive or negative case will be used to provide additional information to the LLM. This allows a knowledge base to be integrated into the medical application without exposing additional medical images to third-party LLMs.

To store images as vectors in the knowledge base, the images need to be embedded. With the local models used in this medical application, their infrastructure can be modified to use their classification layer as the output to be vectorised. This allows the X-ray images to be put into a tensor format which can be vectorised. With Covid-Net and DenseNet having high accuracies in detecting Covid-19 from chest X-rays, these models were used as classifiers for two separate knowledge bases. These knowledge bases consist of classification vectors for the X-ray images contained in the training distribution of the CXR-3 dataset. Although VGG also demonstrated high accuracy, using its classification layer resulted in a model that is five times greater in size than the other local models (500+ MB). To prevent the application from running out of memory, VGG was not chosen to vectorise images. 

During the process of analysing LLMs with knowledge bases, the GPT-4.5-Preview was removed from the OpenAI API. Although this restricts the analysis that can be done with this model, it is still included in the results of this research to provide insight into its accuracy and environmental impact.

\section{Results and discussion}
\label{sec:result}

\subsection{Performance}

\begin{table} [htbp]
\caption{Performance of models when tested on Covidx CXR-3 dataset.}
\begin{tabular}{ |p{2cm}|p{1.55cm}|p{1.9cm}|p{2.2cm}|p{1.8cm}|p{3.4cm}|}
\hline
& Accuracy (\%) & Median time taken ($ms$) & Median carbon footprint ($mgCO_2eq$) & Total memory used (\%) & Median carbon footprint per cloud memory used ($mgCO_2eq/MB$) \\
\hline
\multicolumn{6}{|c|}{\cellcolor{blue!10}Local models} \\
\hline
Covid-Net & \textbf{95.5} & 907 & 0.668 & 42.5 & 0.284  \\
\hline
DenseNet & 91.8 & 264 & 0.194 & 40.5 & 0.0788 \\
\hline
ResNet & 82.3 & \textbf{174} & \textbf{0.141} & 41.6 & \textbf{0.0588} \\
\hline
VGG & 93.8 & 221 & 0.174 & 39.7 & 0.0693\\
\hline
\multicolumn{6}{|c|}{\cellcolor{blue!10}LLMs} \\
\hline
GPT-4.5-Preview & 52.3 &  7270 & 974 & \textbf{36.8} & 359 \\
\hline
o4-Mini & 47.5 & 5280 & 543  & \textbf{36.8} & 200 \\ 
\hline
GPT-4.1-Nano & 52.8 & 1670 & 59.4 & \textbf{36.8} & 21.9 \\
\hline
GenAI & 48.5 & 1580 & 220 & 37.9 & 85.5 \\
\hline
\multicolumn{6}{|c|}{\cellcolor{blue!10}LLMs with DenseNet knowledge base} \\
\hline
o4-Mini & 60.5 & 5440  & 548 & 41.2 & 226 \\
\hline
GPT-4.1-Nano & 79.8 & 1730 & 56.4 & 41.2 & 23.3 \\
\hline
GenAI & 73.5 & 1890 & 236 & 42.4 & 100 \\
\hline
\multicolumn{6}{|c|}{\cellcolor{blue!10}LLMs with Covid-Net knowledge base} \\
\hline
o4-Mini & 51.8 & 6460 & 565 & 43.3 & 245 \\
\hline
GPT-4.1-Nano & 79.3 & 3070 & 73.8 & 43.3 & 32.0 \\
\hline
GenAI & 73.8 & 2700 & 249 & 44.5 & 111 \\
\hline
\end{tabular}
\label{table:results}
\end{table}

Apart from Covid-Net, local models show a bias toward diagnosing X-rays as Covid-19 positive. This is shown in Table \ref{table:acc}, where the sensitivity of local models, except Covid-Net, is significantly higher than their specificity. In contrast, LLMs generally demonstrate higher specificity, similar to Covid-Net, regardless of whether a knowledge base is used. This aligns with existing research that found that the Claude-3.5 Sonnet model classified a high number of positive cases as normal (false negative) \cite{Nguyen2024}. Reducing false negatives is important in medical diagnosis as they can risk delayed or missed treatment \cite{Noguchi2023}. Therefore, it can be argued that the bias towards positive diagnosis from most models is safer than allowing false negatives. However, this bias means that these models should be used as supplementary tools rather than automating diagnosis, especially LLMs and ResNet with their lower accuracies.

\begin{table} [htbp]
\caption{Accuracy, specificity, sensitivity, and positive predictive value (PPV) of models.}
\begin{tabular}{ |p{2.8cm}|p{2.5cm}|p{2.8cm}|p{2.8cm}|p{2.2cm}|}
\hline
& Accuracy (\%) & Specificity (\%) & Sensitivity (\%) & PPV (\%) \\
\hline
\multicolumn{5}{|c|}{\cellcolor{blue!10}Local models} \\
\hline
Covid-Net & \textbf{95.5} & \textbf{99.0} & 92.0 & \textbf{98.9} \\
\hline
DenseNet & 91.8 & 84.5 & 99.0 & 86.5 \\
\hline
ResNet & 82.3 & 64.5 &\textbf{100} & 73.8 \\
\hline
VGG & 93.8 & 87.5 & \textbf{100} & 88.9 \\
\hline
\multicolumn{5}{|c|}{\cellcolor{blue!10}LLMs} \\
\hline
GPT-4.5-Preview & 52.3 & 48.0 & 56.5 & 52.1 \\
\hline
o4-Mini & 47.5 & 60.5 & 34.5 & 46.6 \\
\hline
GPT-4.1-Nano & 52.8 & 70.5 & 35.0 & 54.3 \\
\hline
GenAI & 48.5 & 48.5 & 48.5 & 48.5 \\
\hline
\multicolumn{5}{|c|}{\cellcolor{blue!10}LLMs with DenseNet knowledge base} \\
\hline
o4-Mini & 60.5 & 82.5 & 38.5 & 68.8 \\
\hline
GPT-4.1-Nano & 79.8 & 84.5 & 75.0 & 82.9 \\
\hline
GenAI & 73.5 & 96.0 & 51.0 & 92.7 \\ 
\hline
\multicolumn{5}{|c|}{\cellcolor{blue!10}LLMs with Covid-Net knowledge base} \\
\hline
o4-Mini & 51.8 & 78.0 & 25.5 & 53.7 \\
\hline
GPT-4.1-Nano & 79.3 & 83.5 & 75.0 & 82.0 \\
\hline
GenAI & 73.8 & 88.0 & 59.5 & 83.2 \\
\hline
\end{tabular}
\label{table:acc}
\end{table}

Obtaining probabilistic outputs from LLMs proved challenging. This could be safety constraints where the LLMs are trying to prevent the disclosure of sensitive information with the risk that malicious actors may misuse it \cite{Glukhov2023}. Likewise, reluctance to output only probabilities could be because LLMs are better at providing descriptive analysis of X-rays rather than quantifiable probabilities of diseases. To obtain probabilities in the outputs, the input prompt included assurances that the results would be verified by a human radiologist. As evident in Table \ref{table:acc}, LLMs struggle to give accurate probabilities of Covid-19 being present without knowledge bases. This goes against existing work that notes ChatGPT provides accurate X-ray descriptions \cite{Thawakar2025, Ostrovsky2025}. With the scope of the paper restricting the output to not include these descriptions, ChatGPT struggled to give definitive diagnoses to X-rays.

\begin{figure}
    \centering
    \includegraphics[height=2.1in]{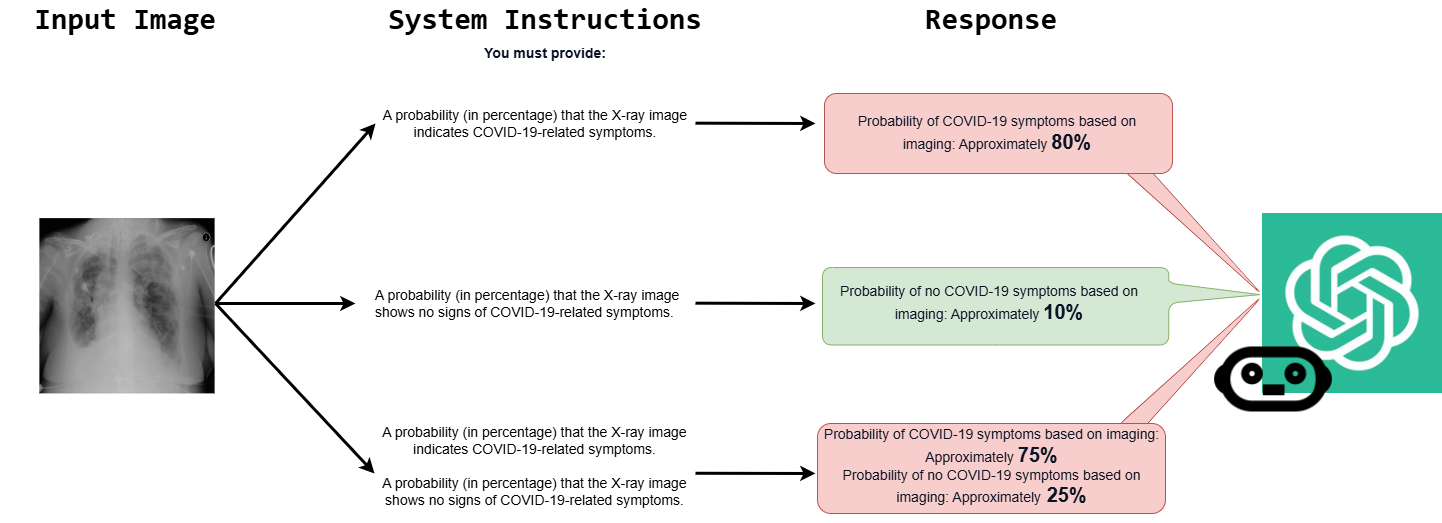}
    \caption{ChatGPT response alteration depending on system instructions.}
    \label{fig:prompt_change}
\end{figure}

LLM outputs also appeared to be biased towards prompt wording. As observed in Figure \ref{fig:prompt_change}, the estimated probability changes with an alteration in the prompt. In cases where API requests were improperly structured and no image was provided, responses aligned to the tone of the prompt; asking for probability of no Covid-19 symptoms pushes a bias that the result is expected to be negative. These results are in line with the suggestions of existing works to use LLM results as a supplementary tool rather than a replacement for patient communication \cite{Tzeng2023, Pedram2024}. So, while LLMs can provide greater detail than smaller models, the risk of bias results means the model should only be used as a supplementary tool, similar to that of the smaller models.

\begin{figure}
    \centering
    \includegraphics[height=3in]{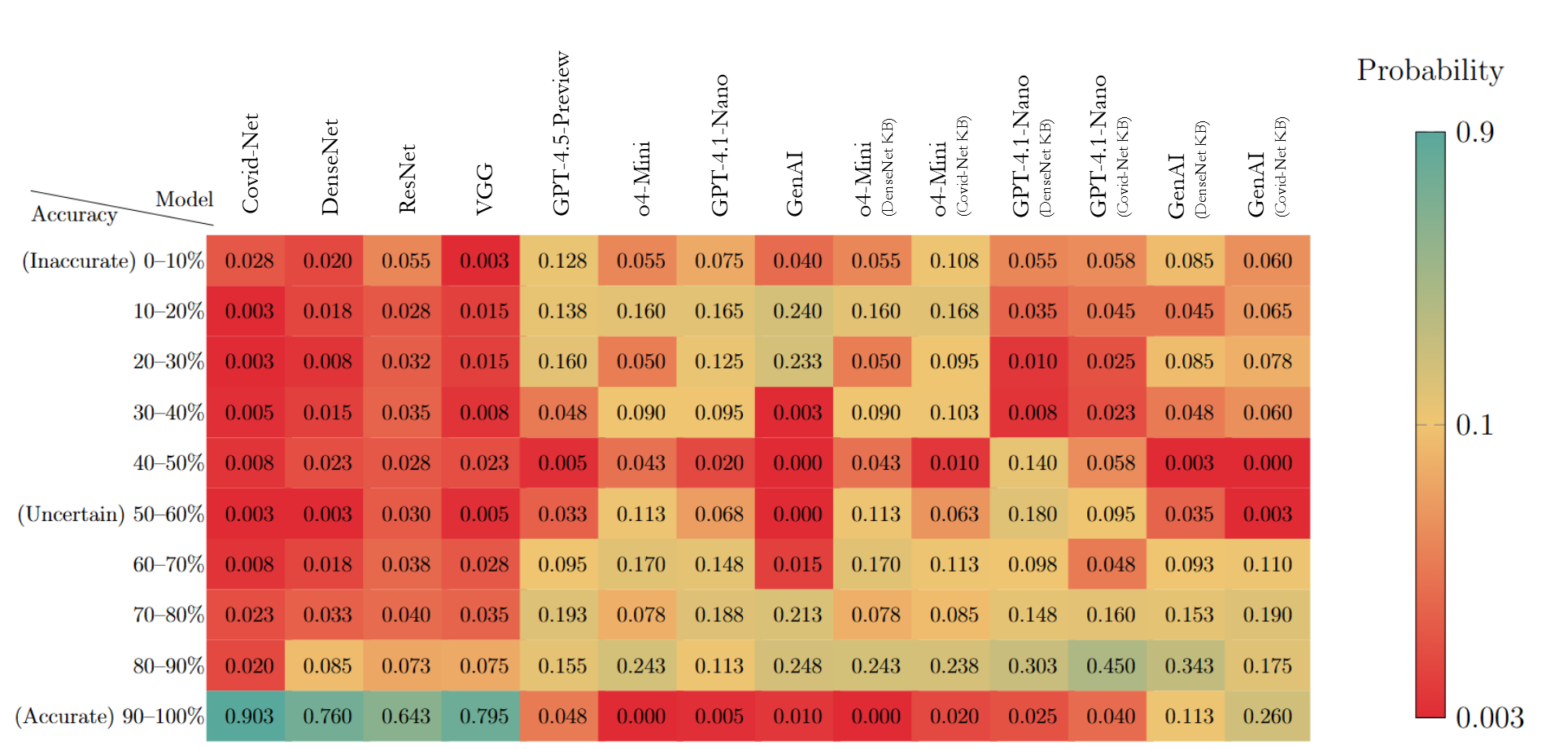}
    \caption{Distribution of model output accuracy, where 100\% shows high accuracy and 0\% shows high inaccuracy (false positive/negative). }
    \label{fig:heatmap}
\end{figure}

Using LLMs without knowledge bases for Covid-19 detection resulted in poor accuracy. As shown in Figure \ref{fig:heatmap}, these models showed high confidence in their output, even when it was a false positive/negative. Meanwhile, they rarely outputted probabilities that suggested uncertainty in the provided X-ray (e.g., approximating an only 51\% probability in an X-ray displaying signs of Covid-19). This highlights the hallucination risks of using LLMs to provide probabilistic outputs for Covid-19, whereby admission of uncertainty is rare.

Introducing knowledge bases generally improved LLM performance and confidence in X-ray analysis, though the extent varied by model and embedding type. GPT-4.1-Nano had a 27\% accuracy increase when using a knowledge base with DenseNet embedding. GenAI saw a 47.5\% increase in specificity with DenseNet embeddings and high-confidence (90–100\%) correct analyses rose from 0.01 to 0.26 with Covid-Net embeddings. Similar results were shown across all LLMs where more confident detections appear when using Covid-Net embedding. However, o4-Mini was influenced the least by the knowledge bases, with the Covid-Net classification knowledge base producing an accuracy still lower than GPT-4.5-Preview had without a knowledge base. This highlights how some models benefit more from knowledge bases \cite{Tayebi2025}. While introducing knowledge bases was found to universally increase LLM accuracy, the type of LLM and the classification model used in the knowledge base influence the impact of the additional information when analysing X-rays.

However, even with knowledge bases, no LLM configuration surpassed Covid-Net in accuracy and confidence in outputs. Over 90\% of Covid-Net's outputs were correctly identified X-rays as positive or negative with a high confidence of 90-100\%. Meanwhile, the other discriminative models only reach this accuracy when increasing the confidence range to 70-100\%. Therefore, while models such as DenseNet and VGG have an accuracy of over 90\%, the confidence in their outputs is less than that of Covid-Net, suggesting a higher level of doubt in detection when using these models. This highlights the superiority of Covid-Net in confidently identifying Covid-19 symptoms compared to existing model architecture, which was fine-tuned to work in the scope outlined in this paper.

For detection of Covid-19, false negatives pose a significant risk. Models that are vulnerable to false negatives could delay the identification, isolation, and treatment of cases. LLMs showed consistently low sensitivity, risking positive cases going undetected. Both ResNet and VGG demonstrated excellent sensitivity and therefore minimise false negatives, but their specificity and PPV values could suggest an unacceptable level of over-diagnosis. Covid-Net balances sensitivity and specificity and therefore would work better in assisting Covid-19 diagnoses.

\subsection{Environmental impact}

\begin{figure} [htbp]
    \begin{minipage}[b]{0.48\textwidth}
    \centering
    \includegraphics[height=2in]{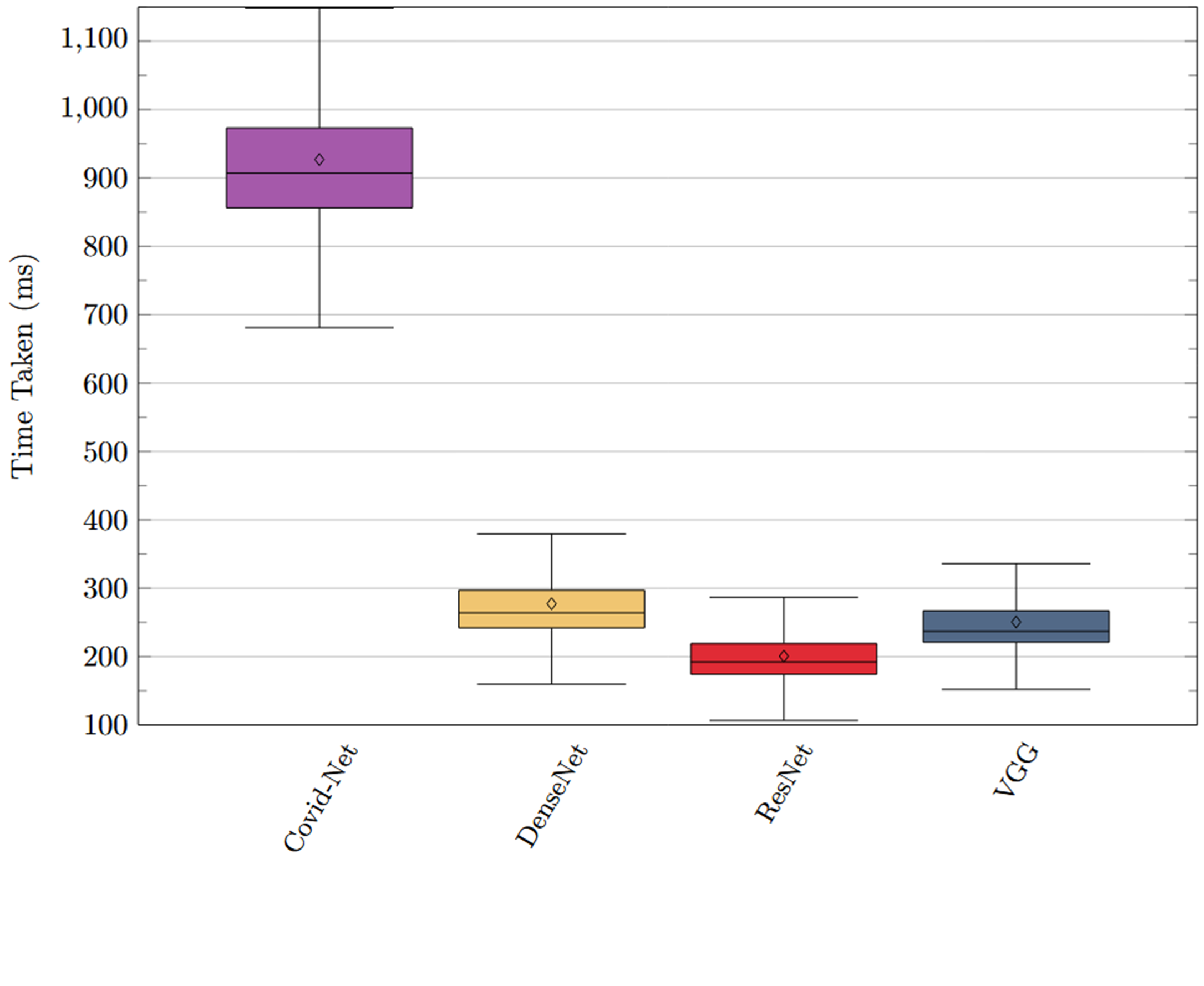}
    \subcaption{local models}
    \label{fig:time_norm}
    \end{minipage}
    \begin{minipage}[b]{0.48\textwidth}
    \centering
    \includegraphics[height=2in]{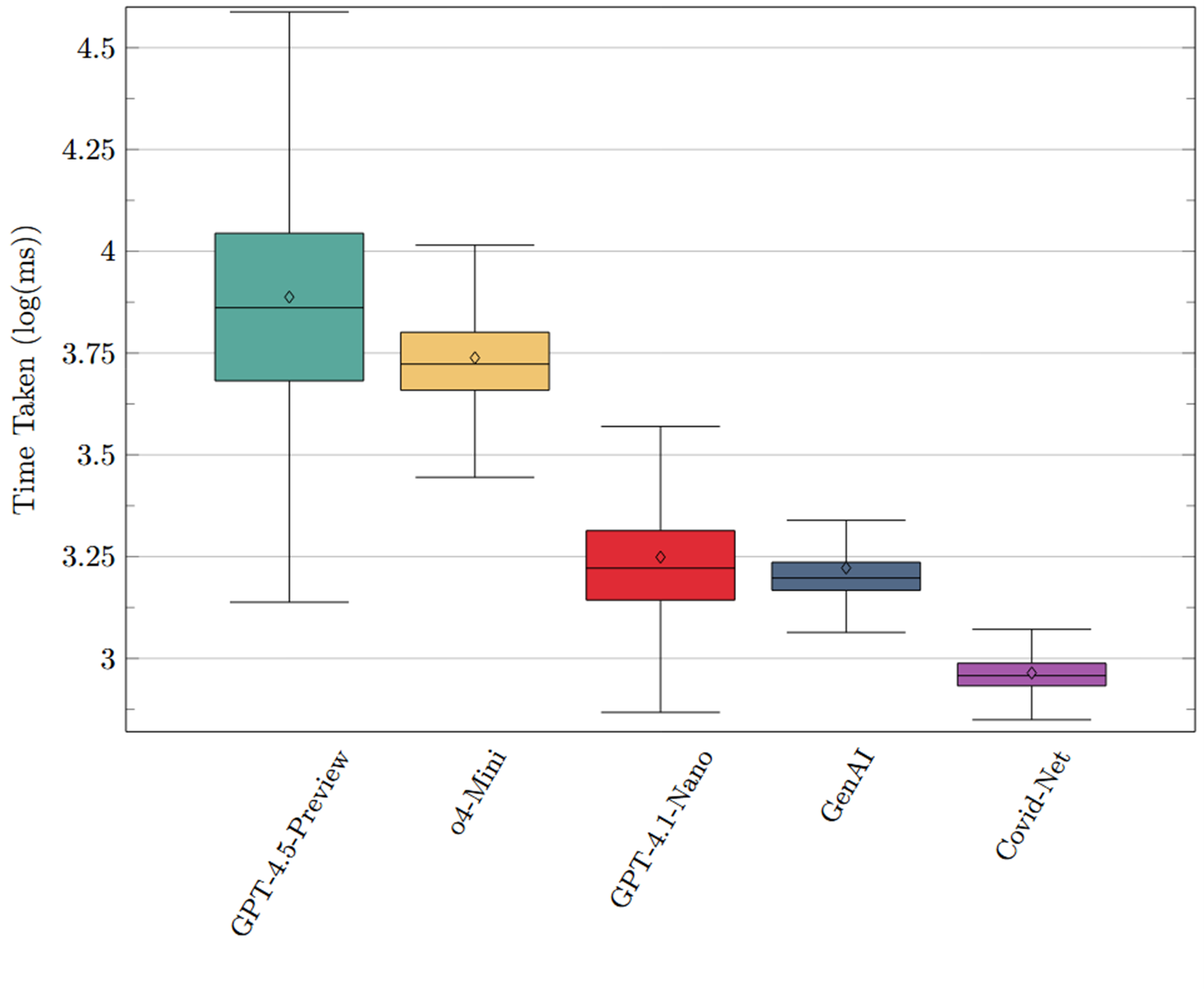}
    \subcaption{LLMs and Covid-Net}
    \label{fig:time_log}
    \end{minipage} %
    \begin{minipage}[b]{0.9\textwidth}
    \centering
    \includegraphics[height=2in]{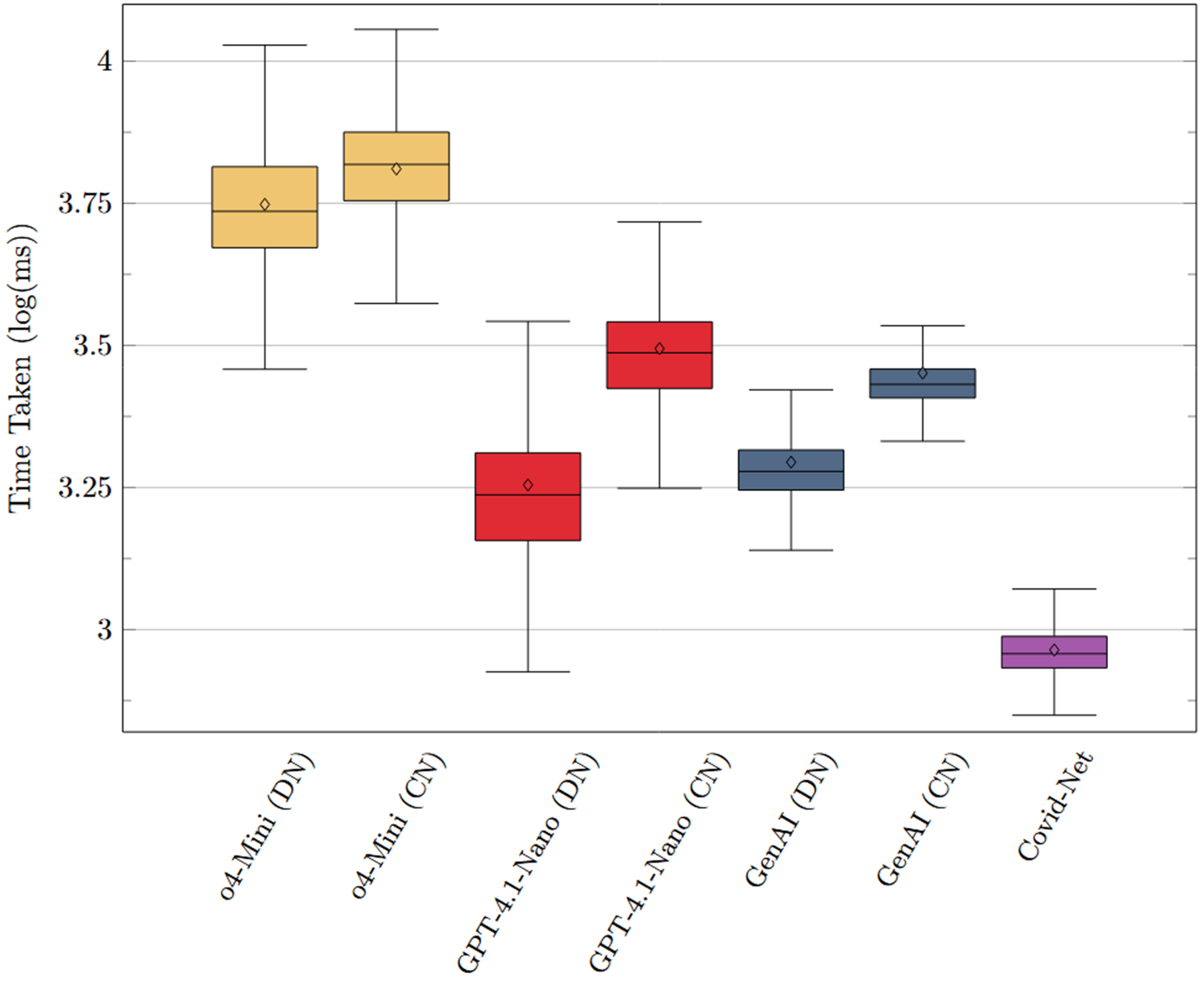}
    \subcaption{LLMs with DenseNet (DN) and Covid-Net (CN) knowledge bases}
    \label{fig:time_log_kb}
    \end{minipage} %
  \caption{Comparing (a) time taken (b), (c) log time taken for models to return output from an X-ray image input.}
  \label{fig:time_box}
\end{figure}

\begin{figure} [htbp]
    \centering
    \includegraphics[height=4in]{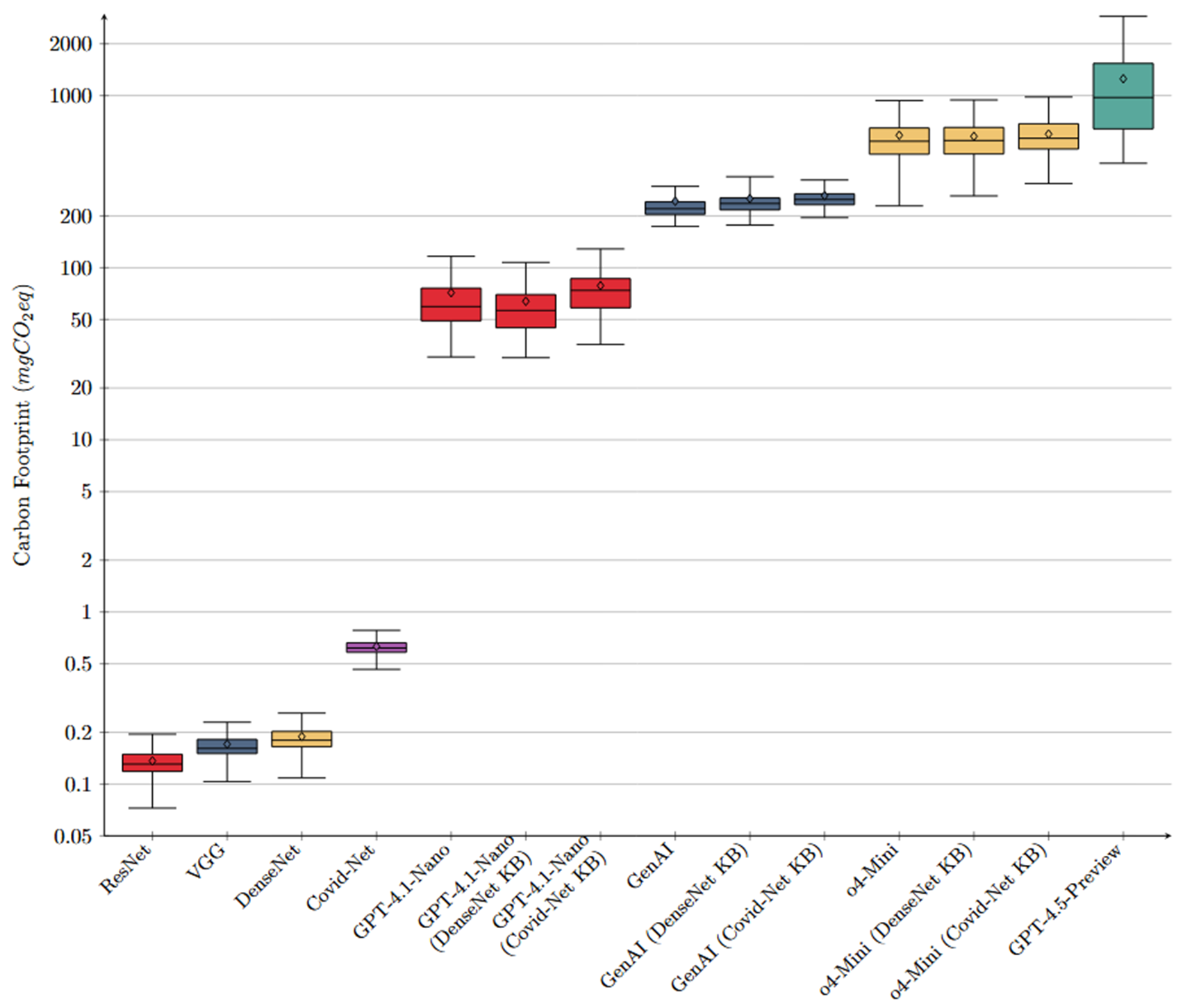}
  \caption{Comparing carbon footprint produced by models to return output from an X-ray image input.}
  \label{fig:carbon_box}
\end{figure}

\begin{figure} [htbp]
  \centering
  \includegraphics[height=2.8in]{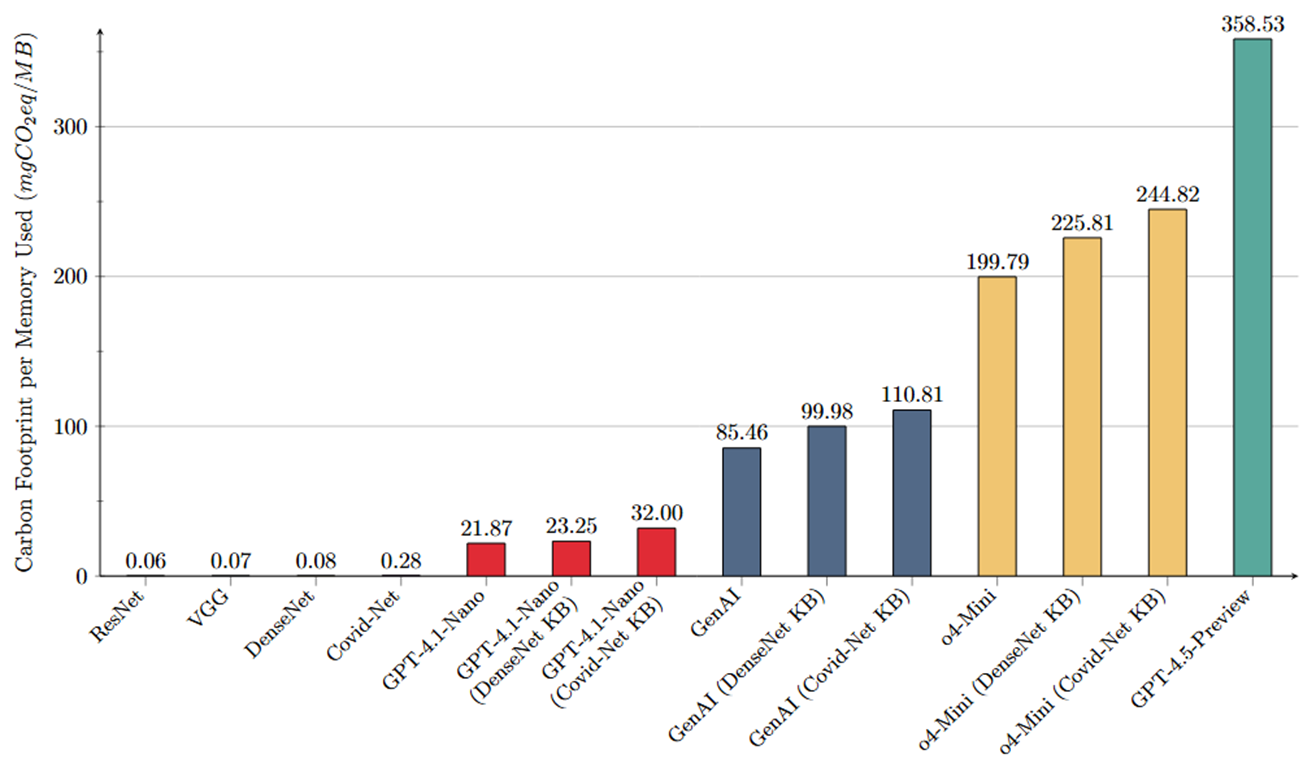}
  \caption{Comparing carbon footprint per memory used produced by models to return output from an X-ray image input.}
  \label{fig:mem_box}
\end{figure}

Among locally deployed models, the carbon footprint of Covid-Net is the biggest. As shown in Figures \ref{fig:time_norm} and \ref{fig:carbon_box}, the efficiency of Covid-Net is around 3.5 times less than that of the other local models. However, Covid-Net’s carbon footprint remained several orders of magnitude lower than those associated with LLM inference. Figure \ref{fig:time_log} shows that the average time taken for GPT-4.5-Preview is 8 times greater than using Covid-Net. Even when taking into consideration the reduced application size of using API calls to LLMs, Figure \ref{fig:mem_box} shows GPT-4.5-Preview still results in a greater footprint a thousand times greater than Covid-Net when analysing X-ray images. So, while Covid-Net is environmentally inefficient compared to other local models, it still produces a smaller carbon footprint compared to LLMs.

While Covid-Net demonstrates superior accuracy, it could be argued that the VGG model provides a more competitive environmental solution. Table \ref{table:results} shows that although its accuracy is 1.7\% lower than Covid-Net, it has a reduced carbon footprint of 74\%. Given that this accuracy is still higher than that of existing models \cite{WangLinda2020, Albahli2021}, it could be argued that this accuracy trade-off is acceptable for reduced carbon emissions \cite{Jung2024, Narimani2025}. However, Figure \ref{fig:heatmap} highlights VGG has less confidence in its outputs, and Table \ref{table:acc} shows that its PPV is 10\% lower than that of Covid-Net. Therefore, while VGG significantly reduces the carbon footprint, the higher risk of incorrect diagnosis means this may not be an appropriate trade-off.

GenAI demonstrated the fastest and most consistent speed across the LLMs. When no knowledge base is used, its mean of 1730ms and interquartile range (IQR) of 253ms is not far off the mean and IQR of Covid-Net (when compared to the other LLMs) which were 929ms and 117ms respectively. With faster response time, GenAI achieves a median emissions 80\% lower than GPT-4.5-Preview despite having similar server hardware and GPU utilisation. Although the speed of GenAI is close to local models, Figure \ref{fig:carbon_box} shows a significant disparity in their carbon footprint produced. So, even when LLM response times are fast, the interaction with a server hosting the LLM results in a significant increase in the carbon footprint, highlighting the adverse effect on energy consumption when using large models for the classification task.

In contrast to locally deployed models, all LLM configurations produced substantially higher carbon footprints. A reason for this energy consumption is that models that generate new text are more carbon intensive \cite{Luccioni2024}. Unlike the small discriminative models that output fixed numeric values, LLMs are generating new text. Therefore, even when the response format of LLMs is restricted to probabilities, this does not eliminate the energy demands of generating new text. This shows that the type of model and the way it formats tasks has a significant role in the carbon footprint produced for an application utilising that model.

Introducing knowledge bases saw a mixed impact in carbon footprint. GPT-4.1-Nano saw a 24.2\% increase when using the Covid-Net embedded knowledge base. Meanwhile, o4-Mini and GenAI only saw a 4.0\% 13.1\% increase in carbon footprints respectively with this knowledge base. This could be because GPT-4.1-Nano has a lower carbon footprint than the other LLMs, meaning a change in its carbon footprint is more noticeable. Meanwhile, using the DenseNet embedded knowledge base saw a decrease of 5.0\% in carbon footprint for GPT-4.1-Nano; while the process of uploading an image and waiting for an output was longer, the time for the LLM running was shorter, which is where the majority of the carbon footprint is being produced. The o4-Mini and GenAI models saw increases of 1\% and 7.2\% respectively with the DenseNet embedded knowledge base. This shows that the environmental impact of using knowledge bases with LLMs varies across the different models and how information is embedded.

\section{Future work and limitation}
With uncertainties surrounding the energy consumption when training models, the calculated carbon footprint produced by the application was constrained to inference with the models. This is a considerable limitation of this paper, whereby the environmental sustainability of training LLMs and local models is not tested. It is recommended that future work investigate the training stage of the models employed in X-ray analysis and whether the collaborative nature of third-party LLMs could demonstrate greater environmental sustainability than that implied through the model inference of the medical application in this paper.

At the time of analysis of the models, there was no suitable way of fine-tuning ChatGPT models with images, nor a way to use ChatGPT extensions within Mendix. This gave ChatGPT a disadvantage in analysis, whereby it may have never been trained on the training dataset the local models were exposed to. Future work should explore the capabilities of fine-tuning LLMs like ChatGPT and analysing their accuracy to local models.

The scope of this paper focused on classifying chest X-rays as either Covid-19 positive or negative. This binary classification allowed for straightforward comparison between models and a clear impact on false negatives, which would typically be a challenge due to generative and discriminative models producing different outputs. Consequently, this research does not provide analysis on wider, multi-class scenarios such as identifying different types of diseases from chest X-rays. Future work should determine whether the superior accuracy and environmental efficiency of local discriminative models in binary classification can also outperform generative models in multi-classification scenarios.

\section{Conclusion}
\label{sec:conclusion}

Medical applications have benefitted from AI tools, but their diagnostic accuracy and environmental impact require careful evaluation. With the versatility of LLMs, generative models can be used in classification tasks despite showing a reduced comprehension in these scenarios \cite{Hanzi2024}. In contrast, small discriminative models can outperform the accuracy of LLMs when used in case-specific applications \cite{Hassid2024}. However, reducing model size to minimise environmental impact can adversely affect accuracy. This paper investigates the performance of LLMs and discriminative models in detecting Covid-19 in chest X-rays. ResNet and VGG resulted in the application having a lower carbon footprint, but their lower PPV values highlighted a strong bias to false positives which could lead to over-diagnosis. Although Covid-Net results in a carbon footprint 3.5 times greater than these models, its footprint is still a 99.93\% reduction of GPT-4.5-Preview's. Even more environmentally friendly LLMs such as GPT-4.1-Nano, which has a 94.2\% lower carbon footprint than GPT-4.5-Preview, still produce a footprint 8350\% greater than Covid-Net. Moreover, Covid-Net had an accuracy 15.7\% higher than that of the best performing LLM (GPT-4.1-Nano with DenseNet embedded knowledge base). This highlights the risk of hallucinations when using LLMs for medical diagnosis \cite{Bera2024}. These results show that larger models used in a classification task for Covid-19 detection have a disproportionate carbon footprint and sub-optimal accuracy when compared to small discriminative model. Additionally, discriminative model sizes should not be reduced to an absolute minimum in an attempt to sacrifice accuracy for the lowest possible carbon footprint, especially since their increased carbon footprint is negligible when compared to the power consumption of LLMs. Future work should investigate the carbon footprint associated with training these models used in medical applications and whether discriminative models can continue outperforming generative models in multi-class scenarios.

\bibliographystyle{IEEEtran}
\bibliography{References} 

\end{document}